\documentclass[letterpaper, 10 pt, conference]{ieeeconf}  

\IEEEoverridecommandlockouts                              

\usepackage{amsmath}
\usepackage{bm}
\usepackage{hyperref}
\usepackage{cleveref}
\crefformat{figure}{#2Fig.~#1#3}
\crefformat{equation}{#2Eq.~#1#3}
\usepackage{cite}
\usepackage[utf8]{inputenc}
\usepackage[english]{babel}
\usepackage{graphicx} 
\usepackage{siunitx}
\usepackage{setspace}

\title{\LARGE \bf
Prismatic Soft Actuator Augments the Workspace \\ of Soft Continuum Robots}

\author{Philipp Wand$^{1*}$, Oliver Fischer$^{1*}$, Robert K. Katzschmann$^{1}$
\thanks{*The ﬁrst two authors contributed equally to this work.}
\thanks{$^{1}$ Soft Robotics Lab, ETH Zurich, Switzerland}
\thanks{{\tt\footnotesize \{\href{mailto:wandp@ethz.ch}{wandp},\href{mailto:olivefi@ethz.ch}{olivefi},\href{mailto:rkk@ethz.ch}{rkk}\}@ethz.ch}}}

\begin{document}
\maketitle
\thispagestyle{empty}
\pagestyle{empty}

\begin{abstract}
\onehalfspacing
Soft robots are promising for manipulation tasks thanks to their compliance, safety, and high degree of freedom. However, the commonly used bidirectional continuum segment design means soft robotic manipulators only function in a limited hemispherical workspace.
This work increases a soft robotic arm's workspace by designing, fabricating, and controlling an additional soft prismatic actuator at the base of the soft arm. This actuator consists of pneumatic artificial muscles and a piston, making the actuator back-driveable. We increase the task space volume by 116\%, and we are now able to perform manipulation tasks that were previously impossible for soft robots, such as picking and placing objects at different positions on a surface and grabbing an object out of a container.
By combining a soft robotic arm with a prismatic joint, we greatly increase the usability of soft robots for object manipulation. This work promotes the use of integrated and modular soft robotic systems for practical manipulation applications in human-centered environments. 

\end{abstract}

\singlespacing

\section{Introduction}
Soft robots have interesting mechanical properties. For example, they are able to  bend continuously, and they are more compliant than conventional rigidly linked robots~\cite{rus2015design,polygerinos2017soft,kim2013soft}. Soft robots are able to interact safely with humans and their surroundings, both when they are operating successfully and when they fail\cite{miriyev2017soft, lee2017review}. These properties are useful for manipulation. Another useful property for manipulation is being able to reach various different positions, or reaching the same position with different configurations. For soft manipulators to become viable for general manipulation applications, we must ensure that their workspace enables them to reach different positions.
The most common design for soft manipulators is a coaxial continuum segment setup in which multiple actuated continuum segments are aligned along the same axis. To our knowledge, all previous soft manipulators have used this design, such as manipulators used in works that investigate soft manipulator kinematics~\cite{neppalli2007design,jiang2016honeycomb,toshimitsu2021sopra}, dynamics~\cite{marchese2016dynamics,katzschmann2019dynamic}, control~\cite{falkenhahn2015modelbased,katzschmann2015autonomous,santina2020model,kazemipour2021robust,fischer2022dynamic}, and learning~\cite{zhang2017toward,jiang2021hierarchical,gillespie2018neural,bruder2021data}. This design utilizes the continuous bending nature of the segments to reach desired positions. 

\begin{figure}[t]
    \centering
    \includegraphics[width=\columnwidth]{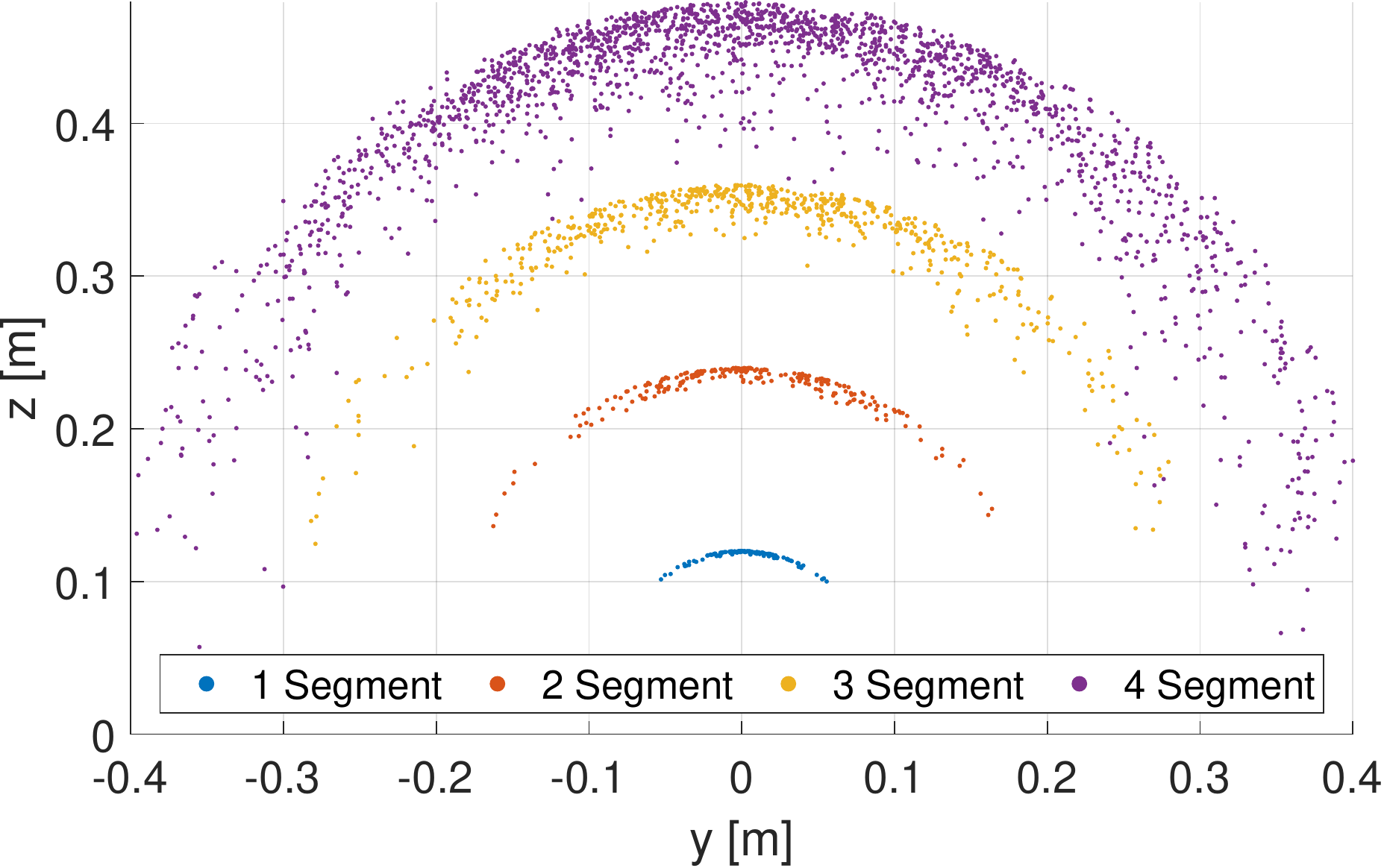}
    \caption{The workspaces of soft continuum manipulators with different numbers of segments. The data was obtained through Monte-Carlo sampling of forward kinematics. Manipulators with more segments have an increased workspace, but retain the same workspace shape.}
    \label{fig:forwardkin}
\end{figure}

However, since all of the segments are coaxial, the workspace can be approximated by a hemisphere with only a slightly variable radius. Crossections of this workspace for different numbers of manipulator segments are shown in \cref{fig:forwardkin}. This workspace restricts soft manipulators' ability to perform manipulation tasks, and it requires a specially designed environment for the manipulator to work properly. These restrictions must be overcome so that we can develop fully compliant and safe soft manipulators that are able to perform general manipulation tasks in a multitude of environments.

One option to overcome these restrictions would be to add more coaxial continuum segments to the manipulator, such as seen with the manipulator used in \emph{Jiang et al.}~\cite{jiang2021hierarchical}, which has 4 continuum segments. However, this approach would not increase workspace significantly, as the coaxiality of the added segments means that they preserve the hemispherical shape of the workspace. This can be seen in \cref{fig:forwardkin}. Additionally, control is a concern: the forward kinematic error of the piecewise constant curvature (PCC) model increases with each added segment. The error of the dynamic model similarly increases with additional segments. The mass of the manipulator also increases with added segments, thus weighing it down. These effects are undesired for dynamic motions. Instead, we could add a prismatic joint to the base of the manipulator, which would allow us to increase or decrease the height of the spherical workspace. This would greatly increase the total workspace while adding only a single degree of freedom, thus keeping model error relatively small. Since the joint would be added at the base, the weight of the manipulator itself would be unaffected, thus preserving its ability to move.

In this work, we design, model, and control a backdrivable soft prismatic joint guided by the following principles:

\textit{Modularity:} The design should be easy to swap, so we can quickly improve on the design and perform experiments. Modularity also allows flexibility in the choice of prismatic actuator; different actuators can be tested. Since the soft arm used in this work is fully pneumatically driven, we additionally constrain the prismatic joint to be pneumatically driven to ease the integration.

\textit{Model Compatibility:} The analytical model should be compatible with common kinematic and dynamic soft manipulator models so that existing controllers can be easily integrated. In particular, it should be compatible with the commonly used piecewise constant curvature (PCC)~\cite{webster2010review} kinematic model and its respective dynamics, which are described by the Euler-Lagrangian equation~\cite{falkenhahn2015modelbased}, or the Augmented Rigid Body~\cite{katzschmann2019dynamic} approach.

\textit{Function Preservation:} The actuator should preserve the attached soft manipulator's functions, particularly its lateral stability and general safety. Therefore, the actuator must be laterally stiff and axially compliant (i.e., it must be backdrivable). 
\subsection{Related Work}
The field of soft robotics offers a variety of actuation methods~\cite{lee2017review,zaidi2021actuation,boyraz2018overview}. These methods can be divided into 3 main categories: tendon-driven actuation, fluidic actuation, and electroactive polymer actuation. Since tendons are not suitable for bidirectional prismatic actuation, and electroactive polymers currently have low strain rates, we chose to focus on fluidic actuation. 

Pneumatic Artificial Muscles (PAM), also known as McKibbens, offer good contraction ratios and are easily manufactured~\cite{Kurumaya2017DesignStructure}. They function by inflating a tube inside an inextensible woven mesh that is able to expand and contract. State of the art McKibbens can create strains of up to 35\%, while other types of PAMs can result in higher strains as presented below (\cite{Yang2019}). \cite{Koizumi2020} shows how McKibbens can be braided if higher contraction ratios are needed.

\emph{Usevitch et al.} developed a bidirectional pneumatic muscle that is driven by a combination of pressure and vacuum~\cite{Usevitch2018APAM}. Their pneumatic muscle is able to achieve excellent contraction ratios of 1000\% by using a multichamber design.

\emph{Kazutoshi et al.} used structure-integrated cylinders to create a compact linear actuator for pneumatic actuation. Their approach is comparatively lightweight and allowed their robot to perform jump-and-hit motions.

\emph{Han et al.} presented a pneumatic muscle that buckles a support structure to achieve higher contraction ratios~\cite{Han2018ARatio}. Their design offers high contraction ratios and can be 3D-printed.

\emph{Jiang et al.} used a soft manipulator to perform basic tasks with kinematic learning-based control~\cite{jiang2021hierarchical}. They mounted the manipulator on a linear rail to open a drawer. While this extended the taskspace horizontally, the soft manipulator's vertical workspace was still restricted. We note that the authors likely use a kinematic controller for this 4-segment manipulator and perform their control in the timeframe of minutes. To the best of our knowledge, dynamic control of a manipulator with this amount of segments has not been shown, likely resulting from the large model errors such a manipulator would have.

\emph{Marchese et al.} designed, modeled and controlled a soft manipulator\cite{marchese2016dynamics}. In one of their experiments, they mounted their manipulator to a slowly moving prismatic base. However, this moving base was neither modeled nor controlled, and it did not play a major part in the experiments. 

In our previous work \emph{Fischer et al.}, we used a 2-segment soft manipulator to perform simple tasks with dynamic model-based control~\cite{fischer2022dynamic}. The limited workspace affected the tasks which were able to be shown. 

\emph{Katzschmann et al.} used an Augmented Rigid Body Model to dynamically control a 3-segment manipulator in curvature space. Their results show that quality of tracking performance decreases with added segments.

The three aforementioned works demonstrate the need for a linear joint to increase the workspace 

\subsection{Contributions}
The main contributions of this work are as follows:
\begin{itemize}
    \item We have combined a piston and PAMs to create a bidirectional pneumatic actuator that minimizes steady state error and is backdriveable.
    \item We integrated a prismatic actuator into a soft continuum manipulator, increasing its workspace by $116\%$.
    \item We demonstrated a soft robotic system using multiple soft actuator types to perform workspace-demanding tasks such as a pick-and-place task and grabbing an object from a container.
\end{itemize}

\section{Design and Fabrication}

\begin{figure}
    \centering
    \includegraphics[width=\columnwidth]{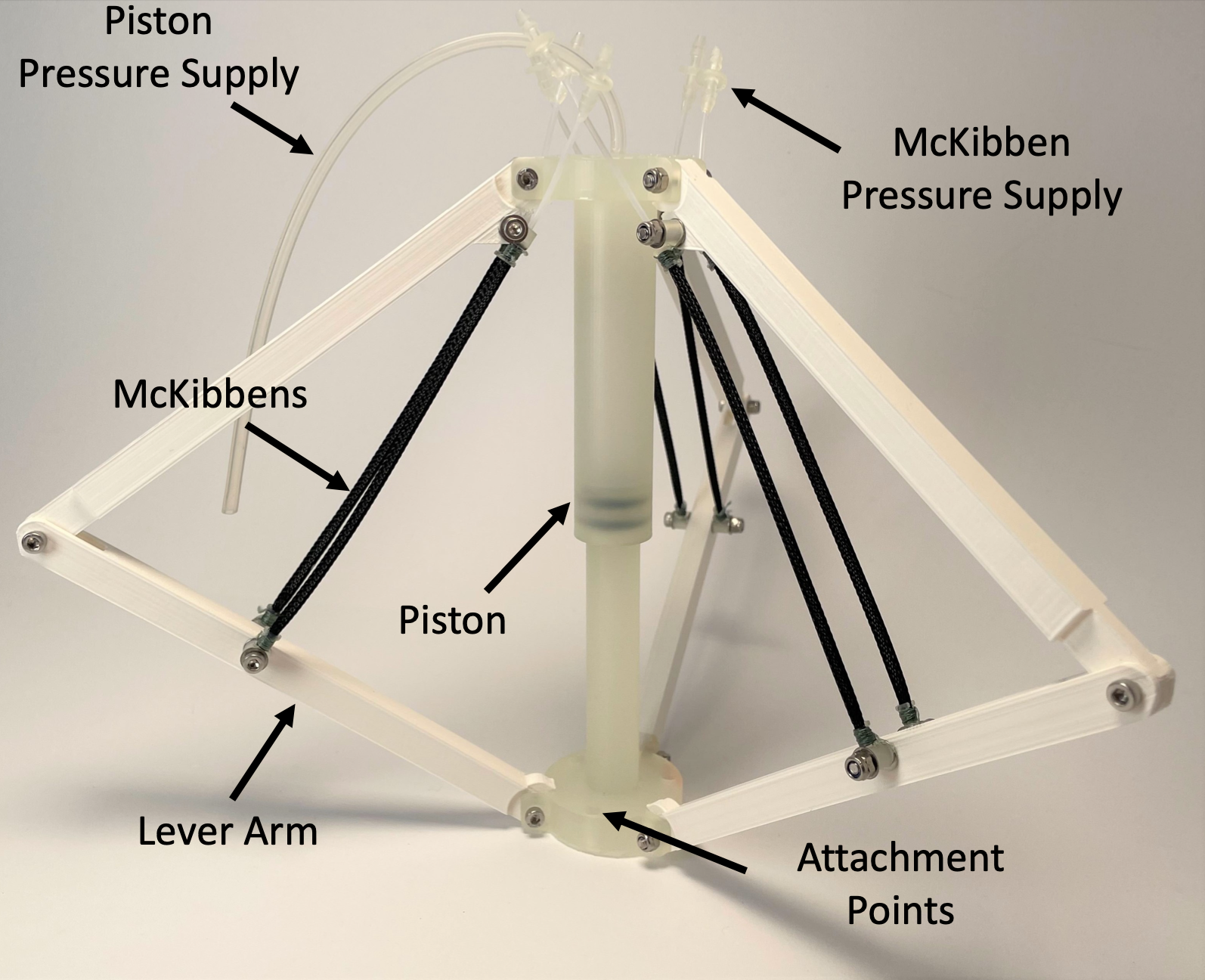}
    \caption{The prismatic soft actuator. The Pneumatic Artificial Muscles (PAMs) of the prismatic actuator are responsible for its  contraction. The central shaft in for of a piston acts as guidance, and also works antagonistically against the PAMs when the piston is pressurized. The lever arms act as mechanical leverage for the PAMs.}
    \label{fig:design}
\end{figure}
\subsection{System Overview and Requirements}
\emph{SoPrA} is a soft continuum manipulator that was developed by \emph{Toshimitsu et al.}~\cite{toshimitsu2021sopra}. It features a fiber-reinforced three-chamber design, and it is pneumatically actuated. We chose to use SoPrA for this work out of convenience, since it is available to our lab.

\emph{SoPrA} consists of two continuum segments that can each actuate in x and y direction, giving it 4 degrees of freedom. The resulting workspace can be seen in \cref{fig:design}. To achieve a maximal increase in the workspace, we choose to place our prismatic joint at the base of the manipulator. Since \emph{SoPrA} can perform quick, dynamic motions\cite{fischer2022dynamic}, we must consider the lateral inertial forces at the base.
\emph{SoPrA} is driven by a pneumatic valve array with pressures up to \SI{2}{bars}. It is not able to create a vacuum. Therefore, the pneumatic actuator must be compatible with this pressure range.
\emph{SoPrA} uses a motion capture system and bend sensors to estimate state. We therefore chose to use motion markers to capture prismatic joint extension.

\subsection{The Prismatic Joint}
Taking this into account, we designed a prismatic actuator consisting of three main components:
\begin{itemize}
    \item A shaft, which guides the actuator as it moves. The shaft creates the required lateral stiffness, and it can use the inbuilt piston to create a downward force.
    \item The pneumatic artifical muscles, which contract upwards and hold the prismatic joint in place.
    \item The lever arms, which are used as contraction multipliers for the PAMs.
\end{itemize}
We choose to actuate our prismatic joint pneumatically because the compressibility of air makes it compliant. Additionally, we are able to use the pre-existing valve array to power the prismatic joint, making integration with the manipulator easy.

The collapsing shaft in the middle of the actuator serves as a backbone to which all other subsystems are attached. We placed symetric mounting platforms on both ends of the shaft. Each of these platforms features three screw clearance holes. We mounted infrared motion capture markers on these platforms for sensing.

The shaft's main active functions are to linearly guide the actuator motion and to provide a downward force using a piston. The actuator must be constrained laterally to prevent oscillation and uncontrolled movements. Therefore, we manufactured the shaft with tight tolerances using stereolithography 3D printing.
To provide the desired extension, the shaft also acts as a piston. We sealed it with two o-rings, and placed a pressure inlet valve at the top. We coated the piston with silicone oil so it can move smoothly while retaining tight tolerances.

PAMs are an ideal choice for the prismatic actuator's contraction mode. While alternative actuators could also be used for this purpose, PAMs are easier to manufacture and control than, for example, soft silicone actuators, which are more suited to bending than prismatic motion. PAMs also have good force-weight ratios, and their pressure-strain relation minimizes drift.
The PAMS were manufactured by embedding a silicone tube into a PET cable sleeve and sealing the ends using a fishing line. Attachment points were directly tied onto the mesh during the sealing process. The bottom of the PAM was completely sealed, and the top featured a small inlet valve to allow for actuation. The PAMs measured \SI{13.2}{cm} when fully relaxed and \SI{10.0}{cm} when fully contracted.

The lever arms act as mechanical leverage for the PAMs. They were 3D-printed, and their aspect ratios were based on the human forearm\cite{tondu1997mckibben}. Using the lever arms, the contraction created by the PAMs increased by 250\%, while the force output decreased by 60\%. Two PAMs were mounted on each lever arm to ensure sufficient force output.

\section{Modeling}
\begin{figure}
    \centering
    \includegraphics[width=0.9\columnwidth]{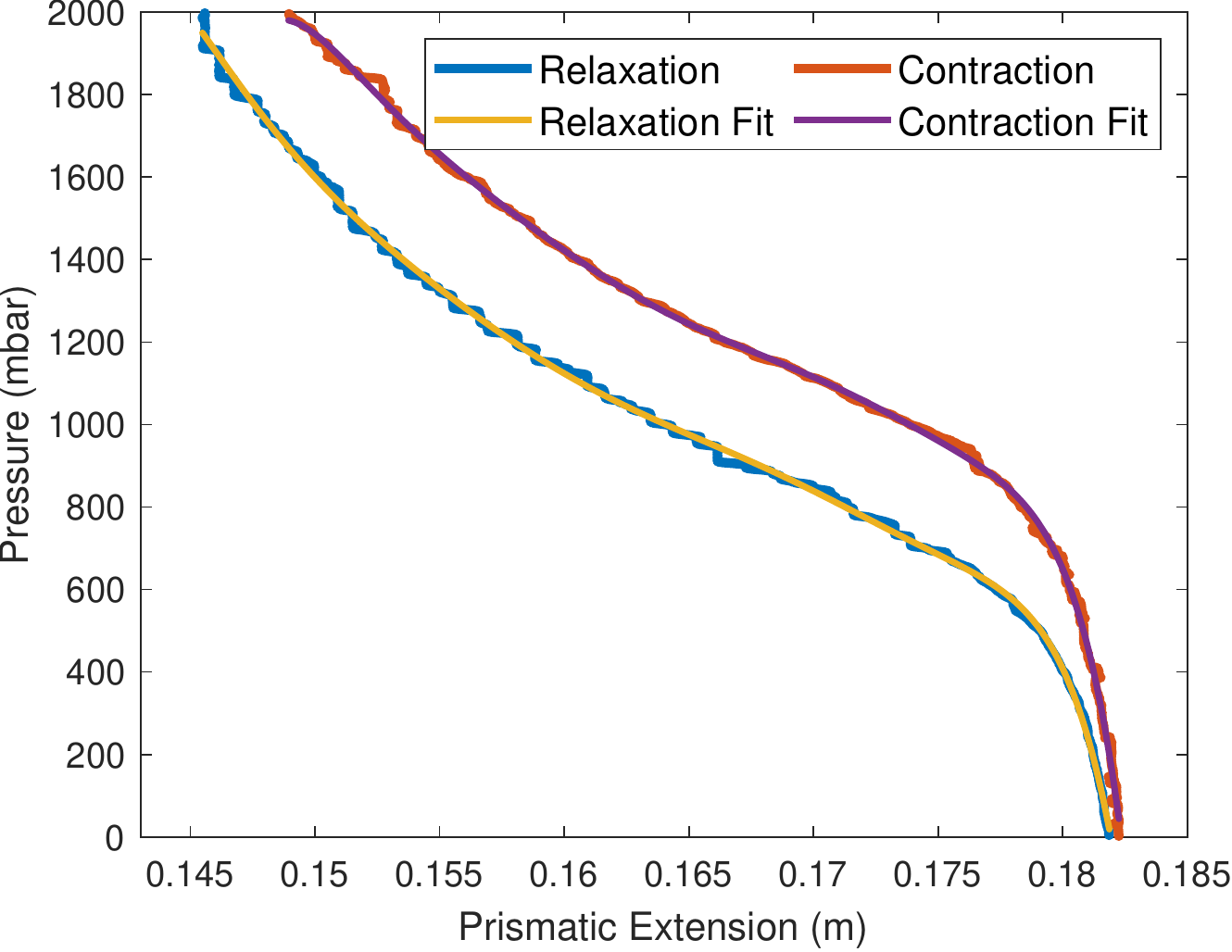}
    \caption{Pressure-extension relationship of the PAMs manifests hysteresis. The two curves represent contraction and relaxation, respectively. We interpolated between the hysteresis functions based on velocity to obtain the local pressure-extension relationship.}
    \label{fig:hysteresis}
\end{figure}
\subsection{The Prismatic Joint}
To integrate the new actuator into the robotic system, we had to model its dynamic behavior. The actuation dynamics of the piston can be described as follows:
\begin{equation}
    \ddot{q}_a = \frac{A \cdot p_p}{m_{tot}}
    \label{eq:pistonactuation}
\end{equation}
where $q$ is the prismatic extension, $A$ is the surface area of the piston, $p_p$ is the piston input pressure, and $m$ is the combined mass of the inner shaft and the attached manipulator. We did not consider friction because the shaft was lubricated.
It is not easy to analytically describe the PAM's dynamics due to the relation between their force and contraction. However, we were able to make some approximations to simplify their dynamics.
We performed experiments to identify the relationship between PAM input pressure and the static position of the PAMs, $p_{s}(q)$. The results can be seen in \cref{fig:hysteresis}. We found that the relation depends on the PAM's actuation mode; during contraction, higher pressures were required to reach an equal prismatic position. This results from the PAM's hysteresis. Therefore, we measured a curve for PAM relaxation $p_{s,r}(q)$ and for PAM contraction $p_{s,c}(q)$. To estimate the true value of $p_{s}(q)$, we used the prismatic velocity $\dot{q}$ to approximate the current actuation mode, and we used a clipping function to interpolate between the two static pressures:
\begin{equation}
    p_{s}(q,\dot{q}) =
    \begin{cases}
        p_{s,r}(q),& \text{if } \dot{q} < \dot{q}_{th,r}\\
        p_{s,m}(q) \cdot (1 + \frac{\dot{q}}{\Delta \dot{q}_{th}}),& \text{if } \dot{q}_{th,r} < \dot{q} < \dot{q}_{th,c}\\
        p_{s,c}(q),& \text{if } \dot{q} > \dot{q}_{th,c}
    \end{cases}
\end{equation}
where $\dot{q}_{th,r}$ and $\dot{q}_{th,c}$ are the threshold velocities for PAM relaxation and contraction, respectively. \mbox{$\Delta \dot{q}_{th} = \dot{q}_{th,r}-\dot{q}_{th,c}$} is the difference between the thresholds. \mbox{$p_{s,m}(q)=\frac{1}{2} (p_{s,r} + p_{s,c})$} is the static pressure between $p_{s,c}$ and $p_{s.r}$.
Therefore, the PAM's actuation dynamics can be described as follows:
\begin{equation}
    \ddot{q}_a = -\frac{n_{Mc} \cdot r \cdot A_{a}\cdot \Delta p_M(q)}{n_L \cdot m_{tot}}
    \label{eq:mckibbenactuation}
\end{equation}
where $n_Mc$ is the number of PAMs, $n_L$ is the number of lever arms, $r$ is the lever arm aspect ratio, $A_a$ is the estimated PAM surface area, and $\Delta p_M(q) = p_M - p_{stat}(q)$ is the difference between static PAM pressure and PAM input pressure. We assumed that there was a constant relation between over/underpressure and the position. We verified this assumption experimentally.
\subsection{Combined System}
To describe the manipulator dynamics, we used the model developed by \emph{Toshimitsu et al., 2021}\cite{toshimitsu2021sopra}. This model was designed for continuum manipulators and uses the \emph{Augmented Rigid Body}\cite{katzschmann2019dynamic} approach. We added the new prismatic joint to the manipulator's Unified Robot Description Format (URDF) model, which is used to derive dynamic terms from augmented rigid joints.

The dynamic model of the fully assembled manipulator is:
\begin{equation}
    \label{eq:dynamicmodel}
    A(\bm{q})\bm{p} + J^T\bm{f} = B(\bm{q})\bm{\ddot{q}} + K(\bm{q})\bm{q} + D\bm{\dot{q}} + \bm{c}(\bm{q},\bm{\dot{q}}) + \bm{g}(\bm{q}) 
\end{equation}

where $\bm{p}$ is the pressure input vector, $\bm{q}$ is the joint configuration vector, $\bm{f}$ is the task space force vector, $A$ is the actuation matrix, $J$ is the Jacobian, $B(\bm{q})$ is the inertia matrix, $K$ is the stiffness matrix, $D$ is the damping matrix, $\bm{c}(\bm{q},\bm{\dot{q}})$ is the coriolis vector, and $\bm{g}(\bm{q})$ is the gravity vector.
We added the prismatic joint to this model by increasing the rank of $\bm{q}$ by one, adding the prismatic joint extension. We increased the matrices in rank accordingly. We derived the coefficients for $A$ from \cref{eq:pistonactuation,eq:mckibbenactuation}. We derived the new $B(\bm{q})$, $\bm{g}(\bm{q})$, and $\bm{c}(\bm{\dot{q}},\bm{q})$ coefficients from the manipulator's mass, position, and velocity. We assumed that both $K$ and $D$ were 0: $K$ has already been modeled with our hysteresis function, and we assumed that $D$ was 0 due to a lack of friction.
\section{Control}
To control the system, we first created a proportional-derivative controller:
\begin{equation}
    \label{eq:pd}
    \ddot{x}_{ref} = \bm{\ddot{x}}_{des} + k_p \cdot (\bm{x}_{des} - \bm{x}) + k_d \cdot (\bm{\dot{x}}_{des} - \bm{\dot{x}})
\end{equation}

where $\bm{x}_{des}$ is the desired position in the task space, $\bm{\ddot{x}}_{ref}$ is the reference acceleration in the task space, $k_p$ is the proportional gain, and $k_d$ is the derivative gain. The gains were made to saturate at a predetermined magnitude. We could then insert $\bm{\ddot{x}}_{ref}$ into an inverse dynamics\cite{nakanishi2008osccomp} formulation:
\begin{equation}
    \label{eq:inversedynamics}
    \bm{\ddot{q}}_{ref} = J^+(\bm{\ddot{x}}_{ref} - \dot{J}\bm{\dot{q}}) + (I - J^+J)\bm{\tau}_{null}
\end{equation}

where $\bm{\tau}_{null}$ is the nullspace input. In this work, we used $\bm{\tau}_{null} = - k_d\bm{\dot{q}}$ to minimize oscillation.

We could then obtain the desired pressure vector by inverting \cref{eq:dynamicmodel}:
\begin{equation}
    \label{eq:invertforpressure}
    \bm{p} = A^+ \big[ B(\bm{q})\bm{\ddot{q}}_{ref} + K(\bm{q})\bm{q} + D\bm{\dot{q}} + \bm{c}(\bm{q},\bm{\dot{q}}) + \bm{g}(\bm{q}) \big] 
\end{equation}

under the assumption that there were no outside forces ($\bm{f} = \bm{0}$).

When actuating the prismatic joint, we used the pressure input to determine which actuator to use. When the prismatic pressure input $p_{pris}$ was positive, i.e., when the manipulator was supposed to move  upwards, we used the PAMs to pull upwards. When the desired pressure was negative, i.e., when the manipulator was supposed to move downwards, we used the piston to push downwards. The PAMs received the static pressure $p_{s}(q,\dot{q})$ as a feedback term rather than the gravity term $\bm{g}(\bm{q})$, since $p_{s}(q,\dot{q})$ more accurately describes the required force required to hold the manipulator in position.

\section{Experimental Validation}
We performed multiple experiments to assess both the qualitative and quantitative effects of the prismatic joint on the full system. \emph{SoPrA} and the attached prismatic joint were mounted inside a motion capture cage. Nine infrared cameras were used to detect manipulator state. The system was connected to a valve array that was capable of outputting \SI{2}{bar}.

\subsection{Workspace Size}
\begin{figure}[!h]
    \centering
    \includegraphics[width=\columnwidth]{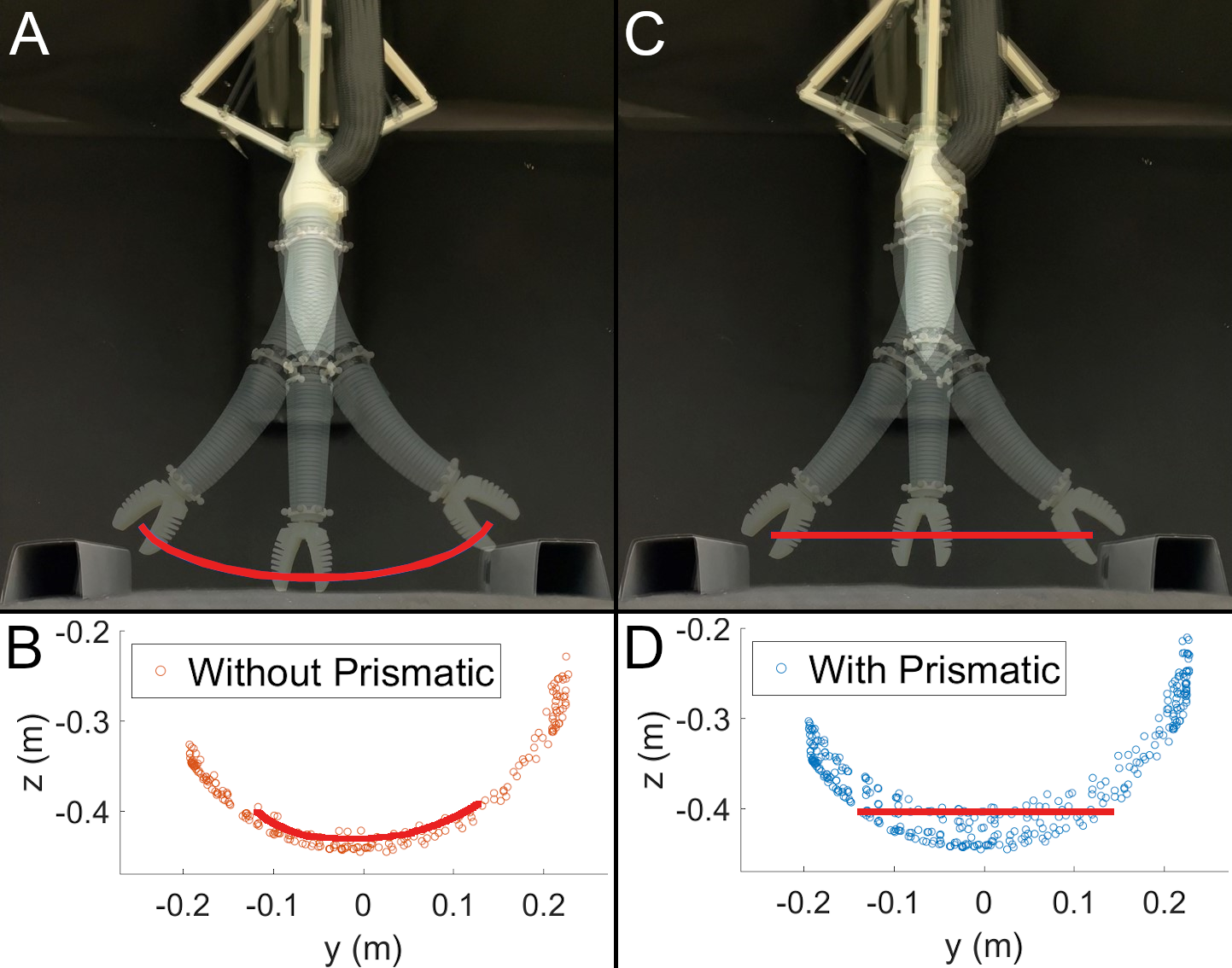}
    \caption{A comparison of the task spaces and possible trajectories when the prismatic joint is used or not. (A) Tracking a straight line without using the prismatic joint. (B) The limited workspace reachable without an additional prismatic joint (C) Tracking a straight line using the prismatic joint. (D) The full workspace reachable due to the additional prismatic joint.}
    \label{fig:workspace}
\end{figure}
We performed Monte Carlo sampling to determine the difference in the workspaces. Slices of the respective workspaces in the $x-z$ direction at $y=0$ can be seen in \cref{fig:workspace}~C,D. We fit hemispherical shells to the respective workspaces to compare volume. The workspace volume when using the prismatic joint was increased by 116\%, or a factor of 2.2. The paths that the manipulator took when it tried to follow a straight line are shown qualitativly in \cref{fig:workspace}~A,B. Without the prismatic joint, the manipulator's workspace cannot accomodate a straight line.

\begin{figure*}[t]
    \centering
    \includegraphics[width=\textwidth]{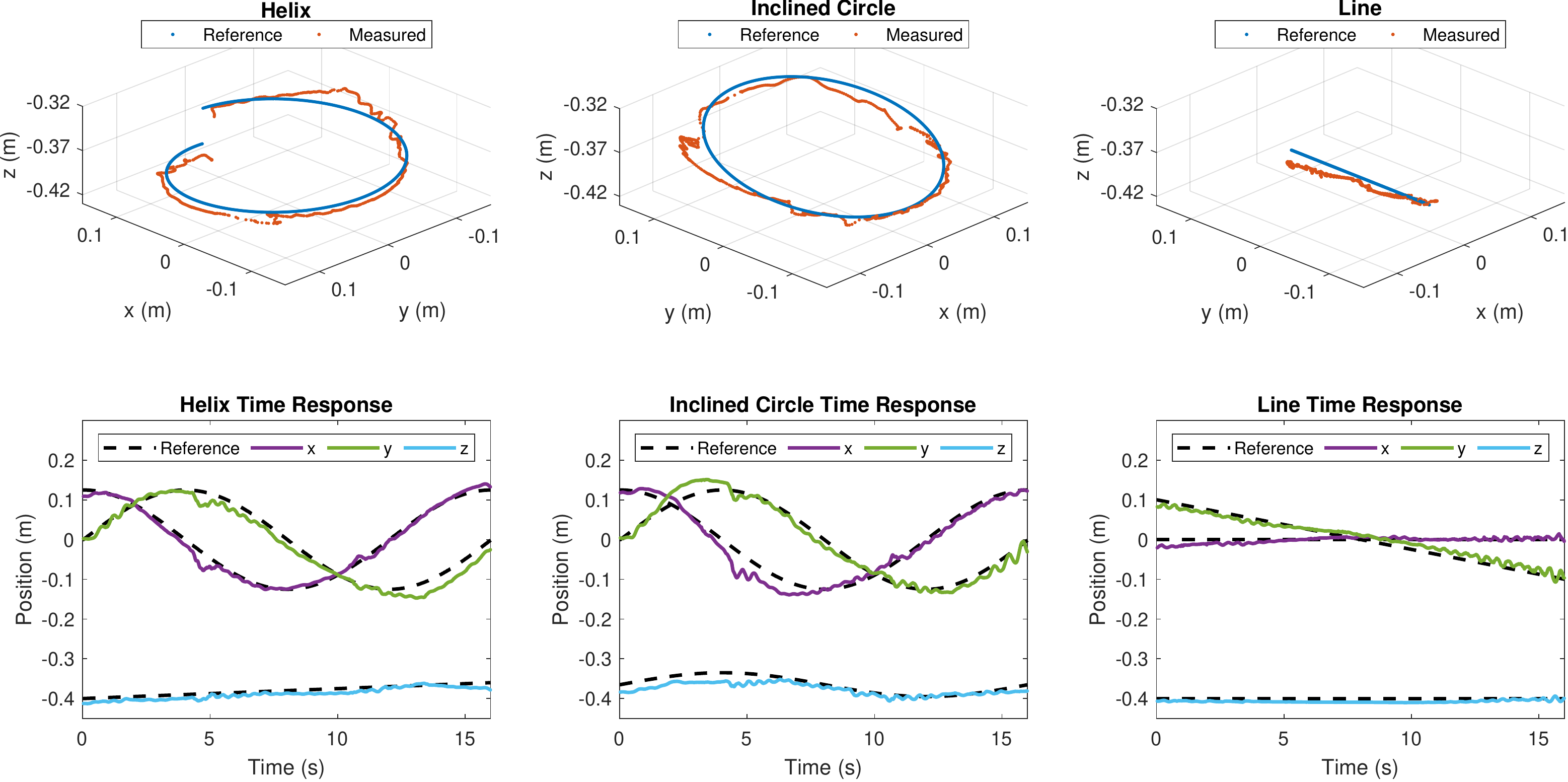}
    \caption{Dynamically tracking a helical, circular, and linear trajectories to verify the new controller's functionality with the integrated prismatic soft actuator. Top: 3D trajectories and their references. Bottom: Time response for the respective trajectories.}
    \label{fig:trajectories}
\end{figure*}

\subsection{Trajectory Tracking}
We investigated multiple dynamic trajectories that had not previously been possible to investigate because of the workspace restriction. The trajectories and their dynamic tracking responses are shown in \cref{fig:trajectories}. The average tip error for the helix, inclined circle and line was \SI{2.1}{cm}, \SI{2.5}{cm}, and \SI{1.6}{cm}, respectively. The positional error of \emph{SoPrA} in previous works was \SI{2.5}{cm} for trajectories within the reachable workspace.

\subsection{Picking Tasks}
\begin{figure*}
    \centering
    \includegraphics[width=\textwidth]{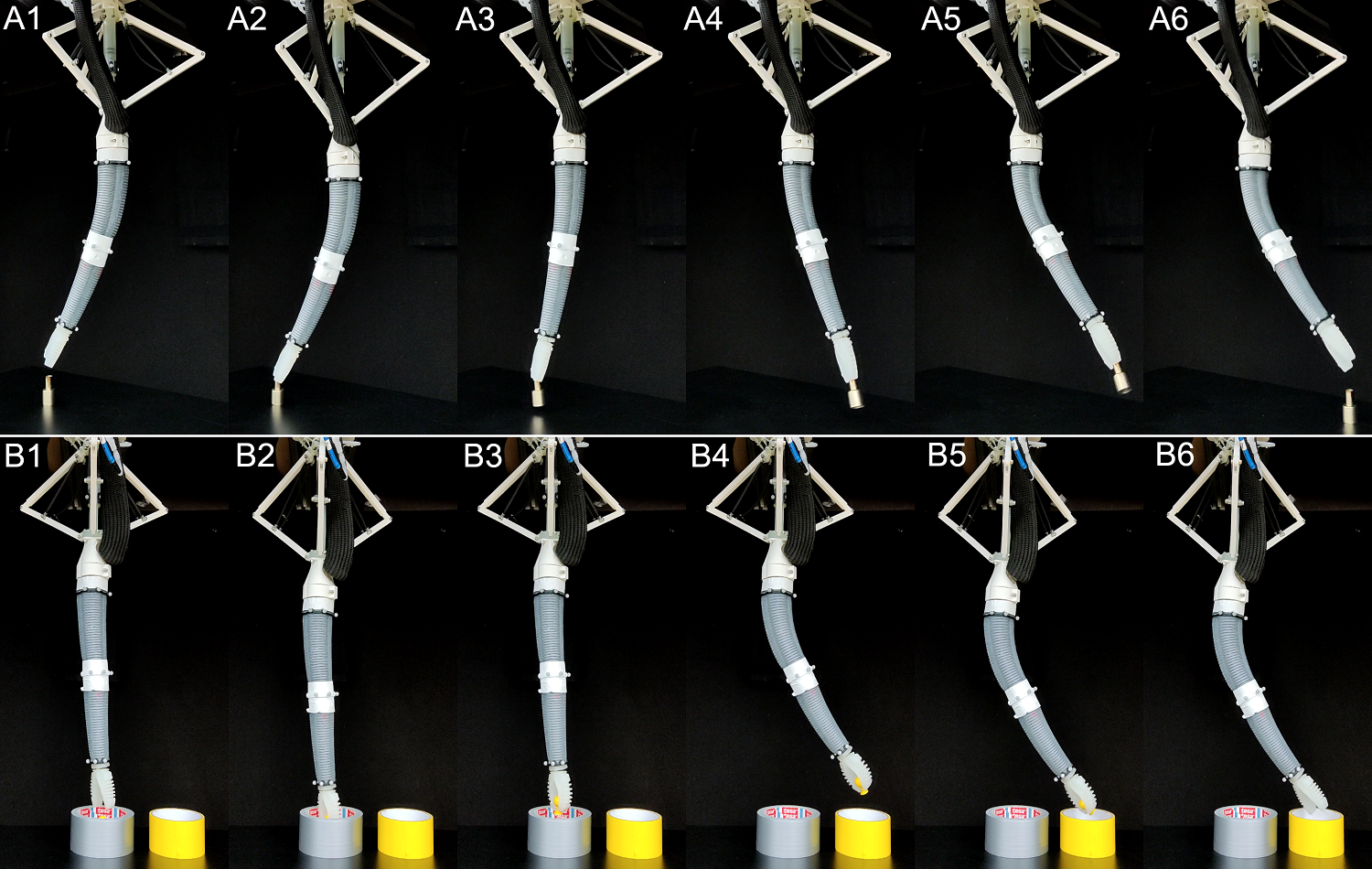}
    \caption{Soft manipulator completing picking tasks that were previously non-reachable. (A1-A6) Picking an object from a surface and dropping it. The frames are spaced in time with 1s intervals. (B1-B6) Picking an object out of a tape roll and dropping it in a different tape roll. The frames are spaced in time with 1s intervals.}
    \label{fig:tasks}
\end{figure*}
We used the extended workspace to perform industry-inspired tasks. The manipulator picked an object from a table, traversed the workspace, and dropped the object on the table in a different position, as seen in \cref{fig:tasks}A1-6. The prismatic joint enabled the manipulator to move down to pick up the object, and to not hit the surface while traversing the workspace. Object location was hardcoded. 

The manipulator also picked an object out of a tape roll and dropped it into a different tape roll, as seen in \cref{fig:tasks}B1-6. The prismatic joint allowed the manipulator to reach inside the tape roll, and to pull the object out of the tape roll. Object and tape roll locations were hardcoded.

Both of these tasks were reproducible with 100\% success rate over 10 runs each.
\subsection{Lateral Stiffness}
\begin{figure}
    \centering
    \includegraphics[width=\columnwidth]{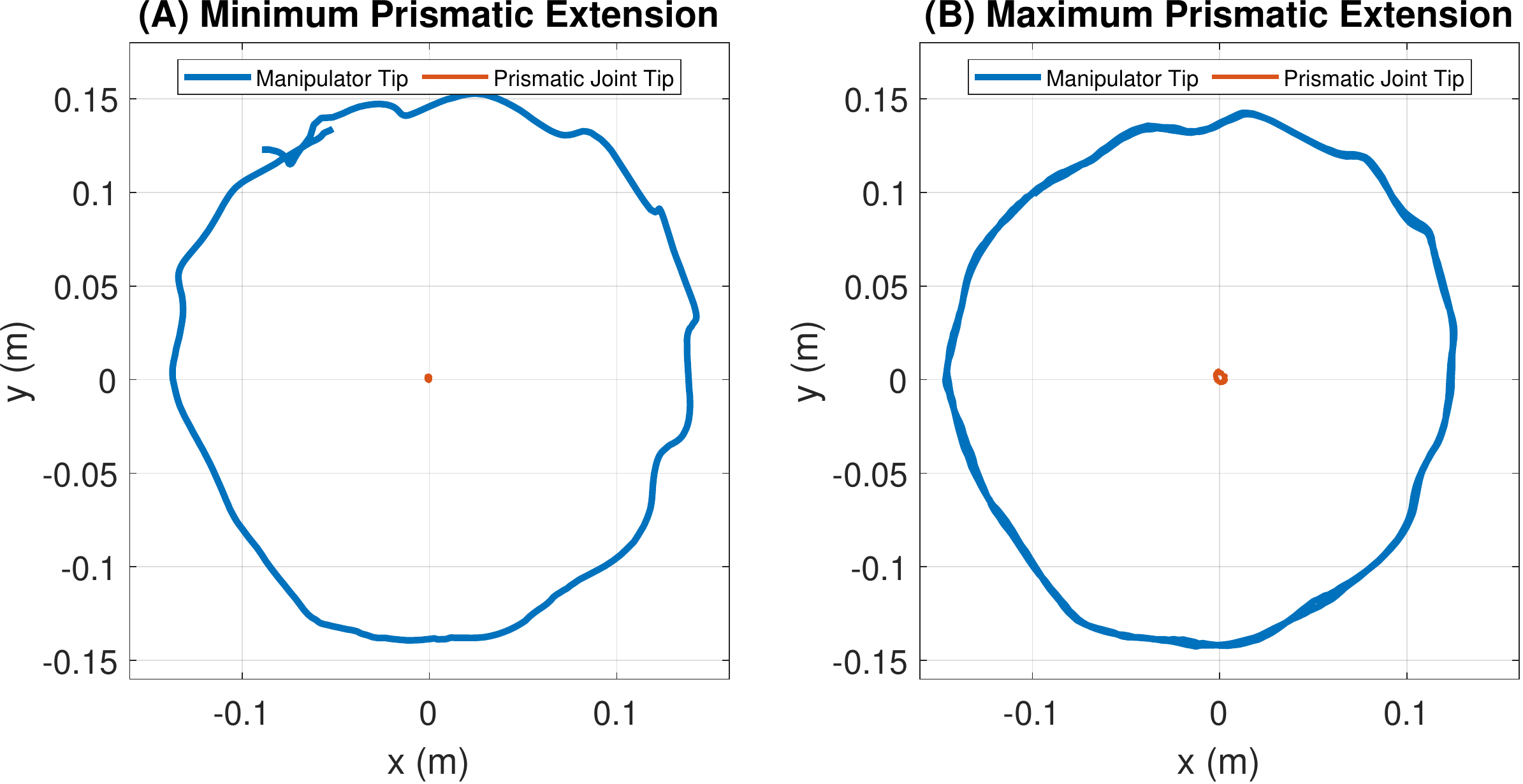}
    \caption{Lateral movement of the manipulator base during operation. (A) Manipulator base movement when the prismatic joint is fully contracted. (B) Manipulator base movement when the base is fully extended.
    \label{fig:lateral}}
\end{figure}
To investigate the system's lateral stiffness, we commanded a circular tip reference trajectory for two cases: when the prismatic joint is fully contracted, and when the prismatic joint is fully extended. The results can be seen in \cref{fig:lateral}. In both cases, the base movement is very small. The base moves slightly more when the joint is fully extended, which is likely because the piston has more play. Furthermore, the path taken by the base is similar to the tip reference trajectory for both cases. A reason for this may be that the base movement is caused by the weight of the manipulator's body.

\subsection{Payload}
We investigated the payload of the prismatic joint by making the manipulator grip a syringe filled with water. This allowed us to control the weight of the gripped object precisely. The maximum payload for the \emph{SoPrA} manipulator without the prismatic joint is \SI{75}{g}. The prismatic joint was able to fully contract while gripping a syringe weighing \SI{75}{g}. We then continued filling the syringe to test the limit of the prismatic joint's payload. The maximum payload we achieved was \SI{200}{g} for full contraction of the prismatic actuator, beyond which the gripper was unable to hold the syringe. 

\section{Discussion}
The addition of the prismatic joint increased both the tracking accuracy and the range of possible trajectories, which can be considered a substantial success for usability in manipulation tasks. Nevertheless, we propose to further decrease the tracking error in future works by improving the dynamic model of both the manipulator and the soft joint. 

The prismatic joint increased the workspace by $116\%$. The value of the increased workspace was shown in multiple tasks. However, a greater increase in the workspace is preferable. Therefore, future works should increase the range of the soft prismatic joint. This could be achieved by using larger PAMs and/or lever arm ratios. 
The piston showed less lateral resistance when it was fully extended. To address this, the inner piston could be elongated so that it is not near the end of the shaft when it is fully extended.

The full system remained compliant when we introduced the soft prismatic actuator. This result is promising: soft manipulators can now perform tasks in a greater task space while retaining the properties that make them interesting for human-robot interaction.

\section{Conclusion}
Our soft prismatic joint design was successfully integrated and precisely controlled as part of a soft robotic manipulation system. The actuator's backdrivability and lateral stability makes its design promising for future robotic applications. The increase in the soft manipulator's workspace allowed us to perform tasks that would be useful in industrial applications. Therefore, we have further closed the gap between soft and rigid manipulators in regards to their ability to perform tasks. In the future, we hope to see soft manipulators replace rigid manipulators in tasks that require human-robot interactions, since soft robots' safety and compliance make them more suitable for these tasks.

\addtolength{\textheight}{-0.5cm}  


\bibliographystyle{IEEEtran}
\bibliography{references}

\begin{thebibliography}{10}
\providecommand{\url}[1]{#1}
\csname url@rmstyle\endcsname
\providecommand{\newblock}{\relax}
\providecommand{\bibinfo}[2]{#2}
\providecommand\BIBentrySTDinterwordspacing{\spaceskip=0pt\relax}
\providecommand\BIBentryALTinterwordstretchfactor{4}
\providecommand\BIBentryALTinterwordspacing{\spaceskip=\fontdimen2\font plus
\BIBentryALTinterwordstretchfactor\fontdimen3\font minus
  \fontdimen4\font\relax}
\providecommand\BIBforeignlanguage[2]{{%
\expandafter\ifx\csname l@#1\endcsname\relax
\typeout{** WARNING: IEEEtran.bst: No hyphenation pattern has been}%
\typeout{** loaded for the language `#1'. Using the pattern for}%
\typeout{** the default language instead.}%
\else
\language=\csname l@#1\endcsname
\fi
#2}}

\bibitem{rus2015design}
D.~Rus and M.~T. Tolley, ``Design, fabrication and control of soft robots,''
  \emph{Nature}, vol. 521, no. 7553, pp. 467--475, 2015.

\bibitem{polygerinos2017soft}
P.~Polygerinos, N.~Correll, S.~A. Morin, B.~Mosadegh, C.~D. Onal, K.~Petersen,
  M.~Cianchetti, M.~T. Tolley, and R.~F. Shepherd, ``Soft robotics: Review of
  fluid-driven intrinsically soft devices; manufacturing, sensing, control, and
  applications in human-robot interaction,'' \emph{Advanced Engineering
  Materials}, vol.~19, no.~12, p. 1700016, 2017.

\bibitem{kim2013soft}
S.~Kim, C.~Laschi, and B.~Trimmer, ``Soft robotics: a bioinspired evolution in
  robotics,'' \emph{Trends in biotechnology}, vol.~31, no.~5, pp. 287--294,
  2013.

\bibitem{miriyev2017soft}
\BIBentryALTinterwordspacing
A.~Miriyev, K.~Stack, and H.~Lipson, ``\BIBforeignlanguage{en}{Soft material
  for soft actuators},'' \emph{\BIBforeignlanguage{en}{Nature Communications}},
  vol.~8, no.~1, p. 596, Sept. 2017. [Online]. Available:
  \url{https://www.nature.com/articles/s41467-017-00685-3}
\BIBentrySTDinterwordspacing

\bibitem{lee2017review}
C.~Lee, M.~Kim, Y.~Kim, N.~Hong, S.~Ryu, and S.~Kim, ``Soft robot review,''
  \emph{International Journal of Control, Automation and Systems}, vol.~15, 01
  2017.

\bibitem{neppalli2007design}
S.~Neppalli and B.~A. Jones, ``Design, construction, and analysis of a
  continuum robot,'' in \emph{2007 IEEE/RSJ International Conference on
  Intelligent Robots and Systems}.\hskip 1em plus 0.5em minus 0.4em\relax IEEE,
  2007, pp. 1503--1507.

\bibitem{jiang2016honeycomb}
H.~Jiang, X.~Liu, X.~Chen, Z.~Wang, Y.~Jin, and X.~Chen, ``Design and
  simulation analysis of a soft manipulator based on honeycomb pneumatic
  networks,'' in \emph{2016 IEEE International Conference on Robotics and
  Biomimetics (ROBIO)}.\hskip 1em plus 0.5em minus 0.4em\relax IEEE, 2016-12.

\bibitem{toshimitsu2021sopra}
Y.~Toshimitsu, K.~W. Wong, T.~Buchner, and R.~Katzschmann, ``Sopra: Fabrication
  \& dynamical modeling of a scalable soft continuum robotic arm with
  integrated proprioceptive sensing,'' \emph{arXiv preprint arXiv:2103.10726},
  2021.

\bibitem{marchese2016dynamics}
A.~D. Marchese, R.~Tedrake, and D.~Rus, ``Dynamics and trajectory optimization
  for a soft spatial fluidic elastomer manipulator,'' \emph{The International
  Journal of Robotics Research}, vol.~35, no.~8, pp. 1000--1019, 2016.

\bibitem{katzschmann2019dynamic}
R.~K. Katzschmann, C.~Della~Santina, Y.~Toshimitsu, A.~Bicchi, and D.~Rus,
  ``Dynamic motion control of multi-segment soft robots using piecewise
  constant curvature matched with an augmented rigid body model,'' in
  \emph{2019 2nd IEEE International Conference on Soft Robotics
  (RoboSoft)}.\hskip 1em plus 0.5em minus 0.4em\relax IEEE, 2019, pp. 454--461.

\bibitem{falkenhahn2015modelbased}
V.~{Falkenhahn}, A.~{Hildebrandt}, R.~{Neumann}, and O.~{Sawodny},
  ``Model-based feedforward position control of constant curvature continuum
  robots using feedback linearization,'' in \emph{2015 IEEE International
  Conference on Robotics and Automation (ICRA)}, 2015, pp. 762--767.

\bibitem{katzschmann2015autonomous}
R.~K. Katzschmann, A.~D. Marchese, and D.~Rus, ``Autonomous object manipulation
  using a soft planar grasping manipulator,'' \emph{Soft Robotics}, vol.~2,
  no.~4, pp. 155--164, 2015.

\bibitem{santina2020model}
C.~Della~Santina, R.~K. Katzschmann, A.~Bicchi, and D.~Rus, ``Model-based
  dynamic feedback control of a planar soft robot: Trajectory tracking and
  interaction with the environment,'' \emph{The International Journal of
  Robotics Research}, vol.~39, no.~4, pp. 490--513, 2020.

\bibitem{kazemipour2021robust}
A.~Kazemipour, O.~Fischer, Y.~Toshimitsu, K.~W. Wong, and R.~K. Katzschmann,
  ``A robust adaptive approach to dynamic control of soft continuum
  manipulators,'' \emph{arXiv preprint arXiv:2109.11388}, 2021.

\bibitem{fischer2022dynamic}
O.~Fischer, Y.~Toshimitsu, A.~Kazemipour, and R.~K. Katzschmann, ``Dynamic
  control of soft manipulators to perform real-world tasks,'' \emph{arXiv
  preprint arXiv:2201.02151}, 2022.

\bibitem{zhang2017toward}
H.~Zhang, R.~Cao, S.~Zilberstein, F.~Wu, and X.~Chen, ``Toward effective soft
  robot control via reinforcement learning,'' in \emph{International Conference
  on Intelligent Robotics and Applications}.\hskip 1em plus 0.5em minus
  0.4em\relax Springer, 2017, pp. 173--184.

\bibitem{jiang2021hierarchical}
H.~Jiang, Z.~Wang, Y.~Jin, X.~Chen, P.~Li, Y.~Gan, S.~Lin, and X.~Chen,
  ``Hierarchical control of soft manipulators towards unstructured
  interactions,'' \emph{The International Journal of Robotics Research},
  vol.~40, no.~1, pp. 411--434, 2021.

\bibitem{gillespie2018neural}
M.~T. {Gillespie}, C.~M. {Best}, E.~C. {Townsend}, D.~{Wingate}, and M.~D.
  {Killpack}, ``Learning nonlinear dynamic models of soft robots for model
  predictive control with neural networks,'' in \emph{2018 IEEE International
  Conference on Soft Robotics (RoboSoft)}, 2018, pp. 39--45.

\bibitem{bruder2021data}
D.~Bruder, X.~Fu, R.~B. Gillespie, C.~D. Remy, and R.~Vasudevan, ``Data-driven
  control of soft robots using koopman operator theory,'' \emph{IEEE
  Transactions on Robotics}, vol.~37, no.~3, pp. 948--961, 2021.

\bibitem{webster2010review}
\BIBentryALTinterwordspacing
I.~Robert J.~Webster and B.~A. Jones, ``Design and kinematic modeling of
  constant curvature continuum robots: A review,'' \emph{The International
  Journal of Robotics Research}, vol.~29, no.~13, pp. 1661--1683, 2010.
  [Online]. Available: \url{https://doi.org/10.1177/0278364910368147}
\BIBentrySTDinterwordspacing

\bibitem{zaidi2021actuation}
S.~Zaidi, M.~Maselli, C.~Laschi, and M.~Cianchetti, ``Actuation technologies
  for soft robot grippers and manipulators: A review,'' \emph{Current Robotics
  Reports}, vol.~2, no.~3, pp. 355--369, 2021.

\bibitem{boyraz2018overview}
P.~Boyraz, G.~Runge, and A.~Raatz, ``An overview of novel actuators for soft
  robotics,'' in \emph{Actuators}, vol.~7, no.~3.\hskip 1em plus 0.5em minus
  0.4em\relax Multidisciplinary Digital Publishing Institute, 2018, p.~48.

\bibitem{Kurumaya2017DesignStructure}
S.~Kurumaya, H.~Nabae, G.~Endo, and K.~Suzumori, ``{Design of thin McKibben
  muscle and multifilament structure},'' \emph{Sensors and Actuators A:
  Physical}, vol. 261, 7 2017.

\bibitem{Yang2019}
H.~D. Yang, B.~T. Greczek, and A.~T. Asbeck, ``Modeling and analysis of a
  high-displacement pneumatic artificial muscle with integrated sensing,''
  \emph{Frontiers in Robotics and AI}, vol.~5, 1 2019.

\bibitem{Koizumi2020}
S.~Koizumi, S.~Kurumaya, H.~Nabae, G.~Endo, and K.~Suzumori, ``Recurrent
  braiding of thin mckibben muscles to overcome their limitation of
  contraction,'' \emph{Soft Robotics}, vol.~7, 4 2020.

\bibitem{Usevitch2018APAM}
N.~S. Usevitch, A.~M. Okamura, and E.~W. Hawkes, ``{APAM: Antagonistic
  Pneumatic Artificial Muscle},'' in \emph{2018 IEEE International Conference
  on Robotics and Automation (ICRA)}.\hskip 1em plus 0.5em minus 0.4em\relax
  IEEE, 5 2018.

\bibitem{Han2018ARatio}
K.~Han, N.-H. Kim, and D.~Shin, ``{A Novel Soft Pneumatic Artificial Muscle
  with High-Contraction Ratio},'' \emph{Soft Robotics}, vol.~5, no.~5, 10 2018.

\bibitem{tondu1997mckibben}
B.~Tondu and P.~Lopez, ``The mckibben muscle and its use in actuating
  robot-arms showing similarities with human arm behaviour,'' \emph{Industrial
  Robot: An International Journal}, 1997.

\bibitem{nakanishi2008osccomp}
J.~Nakanishi, R.~Cory, M.~Mistry, J.~Peters, and S.~Schaal, ``Operational space
  control: A theoretical and empirical comparison,'' \emph{The International
  Journal of Robotics Research}, vol.~27, no.~6, pp. 737--757, 2008.

\end{thebibliography}

\end{document}